\documentclass[11pt, a4paper]{article}

\usepackage{booktabs}
\usepackage{graphicx}
\usepackage{subcaption}
\usepackage{caption}  
\usepackage{siunitx}  
\usepackage{tabularx}
\usepackage{times}
\usepackage{url}
\usepackage{xcolor}
\usepackage{soul}
\usepackage[numbers]{natbib}
\usepackage[final]{neurips_2023}

\newcommand{\imagenetid}[1]{\texttt{\textbf{#1}}}

\title{A Pytorch Reproduction of Masked Generative Image Transformer}
\author{Victor Besnier \\ Valeo.ai, Prague
        \And
        Mickael Chen \\ Valeo.ai, Paris}
\date{}

\begin{document}

\maketitle

\begin{abstract}
In this technical report, we present a reproduction of MaskGIT: Masked Generative Image Transformer \cite{chang2022maskgit}, using PyTorch \cite{paszke2019pytorch}.
The approach involves leveraging a masked bidirectional transformer architecture, enabling image generation with only few steps ($8\sim16$ steps) for $512\times 512$ resolution images, i.e., $\sim$64x faster than an auto-regressive approach. Through rigorous experimentation and optimization, we achieved results that closely align with the findings presented in the original paper. We match the reported FID of 7.32 with our replication and obtain 7.59 with similar hyperparameters on ImageNet at resolution $512\times 512$. Moreover, we improve over the official implementation with some minor hyperparameter tweaking, achieving FID of \textbf{7.26}. At the lower resolution of $256\times 256$ pixels, our reimplementation scores \textbf{6.80}, in comparison to the original paper's 6.18. To promote further research on Masked Generative Models and facilitate their reproducibility, we released our code and pre-trained weights openly at \url{https://github.com/valeoai/MaskGIT-pytorch/}.
\end{abstract}

\begin{figure}[htbp]
    \centering
    \includegraphics[width=0.9\linewidth]{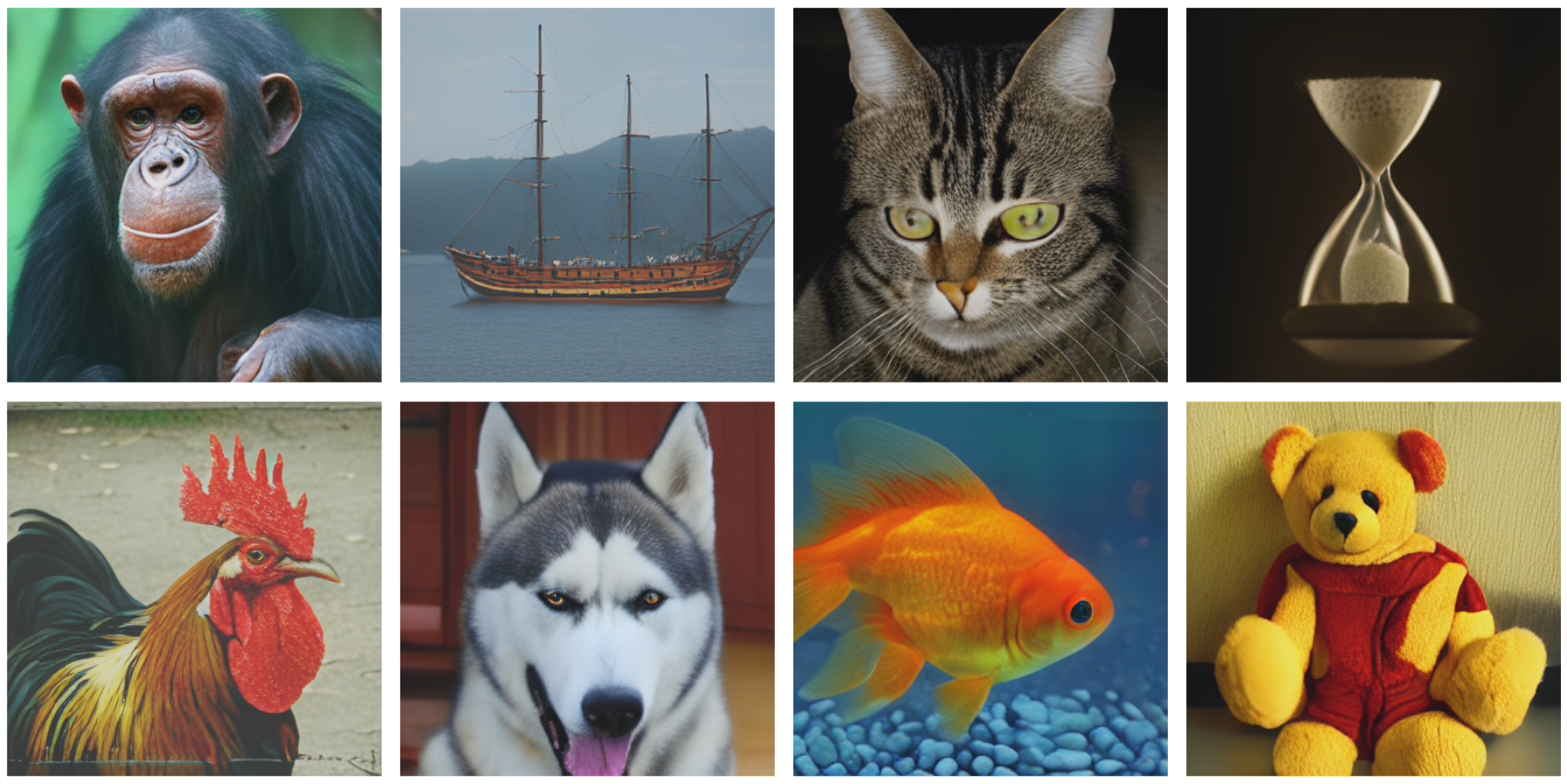}
    \caption{Examples generated on ImageNet at $512\times 512$ demonstrate the effectiveness of our reproduction. Hyperparameters for this set of examples are Gumbel temperature set to 7, Softmax temperature set to 1.3, CFG set to 9, scheduler set to arccos, and scheduler step set to 32}
    \label{fig:qualitative_512}
\end{figure}

\section{Masked Generative Models}
In recent years, advancements in image synthesis have exhibited significant progress. Notably, diffusion models~\cite{dhariwal2021diffusion} have gained considerable traction, surpassing GANs in popularity due to its ease of training and capacity to capture a more diversified distribution. But, this paper directs its focus towards another type of generative methods known as Masked Generative models (MGM) \cite{chang2023muse,chang2022maskgit, li2023mage,villegas2023phenaki,yu2023magvit, yu2023magvit2}. The fundamental concept involves harnessing the capabilities of a VQGAN \cite{esser2020taming} to discretize and reduce an image into visual tokens. Once this discretization is achieved, a bidirectional transformer \cite{devlin2019bert} is trained to unmask a randomly selected set of visual tokens. During inference, the transformer progressively unmasks a fully masked image. This process makes it possible to unmask tokens in parallel. More information can be found in~\cite{chang2022maskgit}.

Conceptually, MGMs have strong connections to several popular and successful deep learning methods. They are very close to Masked Auto-Encoders \cite{he2022mae}, a strong family of models for self-supervised learning, with whom they share similar transformers architectures and the masked reconstruction training objective, except in a token space. MGMs are also connected to auto-regressive models \cite{oord2016pixelrnn,Brown2020Language}, a very strong paradigm for data generation, as both perform generation by iteratively predicting tokens, but only MGMs are able to parallelize the process. Finally, MGMs can be linked to discrete reverse diffusion, when the forward noise process is replaced by a discrete patch dropping scheme. The sampling process is progressively retrieving patches over multiple iterations. As MGMs are in very close proximity to those tried and tested methods, we argue that they are very promising research venues that deserve more attention than they currently have. This idea is further supported by the fact that a Masked Generative Model, MAGVIT-v2 \cite{yu2023magvit2}, reached state-of-the-art results for both text to image and text to video generation.

However, despite its promises, it has to be noted that very few public releases of MGMs have been made available so far. At the time of writing, most prominent papers for image generation do not have full release of code and pretrained weights, if any at all. For instance, the founding paper, MaskGIT \cite{chang2022maskgit}, comes with pretrained model and inference script\footnote{\url{https://github.com/google-research/maskgit/}}, but no training nor evaluation scripts. A text-to-image model, MUSE \cite{chang2023muse}, has no public code release at all.
The MAGVIT \cite{yu2023magvit}, introduced for video generation, can also be used for images and do provide code, but no pretrained weights. For video prediction, TECO \cite{yan2023teco} has both code and pretrained weights, but nothing trained on image datasets, and Phenaki \cite{villegas2023phenaki} has no official code release at all. Moreover, when existing, all these codebases are JAX implementations, and very few resources for Mask Generative Models exist in Pytorch. Still, on-going reproduction efforts of MUSE\footnote{\url{https://github.com/lucidrains/muse-maskgit-pytorch}} and Phenaki\footnote{\url{https://github.com/lucidrains/phenaki-pytorch}} in Pytorch should be noted but they have yet to publicize results or pretrained models. To the best of our knowledge, the only complete image Masked Generative Model released in Pytorch is that of MAGE \cite{li2023mage}, that proposed a non-conditional generative model where MaskGIT accepts class token that allows for conditioning.

In this context, this technical report sets out to reproduce the findings of the original MaskGIT paper~\cite{chang2022maskgit}, as their training code is not available and because no convincing reproduction has been publicly released so far.
With this report, we provide all the code necessary for replicating the results, including training code and our pretrained models.
This work aims at helping researchers to explore the topic of Masked Generative Modeling.

We organize this report as follows: in Section \ref{sec:implem}, we succinctly describe the model, training and sampling techniques, and discuss differences between our reproduction and the details in the paper or official implementation. In Section \ref{sec:results}, we present both quantitative and qualitative results obtained from our networks and, in particular, we ablate the different sampling hyperparameters.

\section{Implementation Details}
\label{sec:implem}

We utilized a pre-trained VQGAN consisting of 72.142M parameters, trained exclusively on the ImageNet dataset.
This VQGAN employs a codebook of 1024 codes and reduces a $256\times 256$ (resp. $512\times 512$) image to a $16\times 16$ (resp. $32\times 32$) token representation.
For additional information, readers can refer to the official repository\footnote{\url{https://github.com/CompVis/taming-transformers}}. 

The bidirectional transformer that generates the visual tokens consists of 174.161M parameters on ImageNet $256\times 256$ and 176.307M parameters on ImageNet $512\times 512$. This sums up to 246.30M and 248.44M respectively, including the VQGAN. Hyperparameters settings are shown in Table \ref{tab:transformer}.

\begin{table}
    \caption{Transformer architecture}
    \label{tab:transformer}
    \begin{tabular}{cccccc}\hline
     \textbf{Hidden Dimension} & \textbf{Codebook Size} & \textbf{Depth} & \textbf{Attention Heads} & \textbf{MLP Size} & \textbf{Dropout} \\ \hline
     768              & 1024          & 24    & 16         & 3072     & 0.1     \\
     \hline
    \end{tabular}
\end{table}

For the inputs, we concatenate the conditional tokens with the visual tokens, i.e., the model takes as input ($16\times 16$) + 1 tokens for ImageNet $256\times 256$, and ($32\times 32$) + 1 tokens for ImageNet $512\times 512$. The model is trained for token unmasking, using cross-entropy loss with label smoothing of 0.1. The optimizer employed is AdamW with a learning rate of \(1e^{-4}\), betas=(0.9, 0.96), and a weight decay of \(1e^{-5}\). We utilize an arccos scheduler for masking during training, regardless of the image resolution. Additionally, we drop 10\% of the conditions for classifier-free guidance. We train our methods on both ImageNet $256\times 256$ and $512\times 512$. This well-known classification dataset consists of 1.2M images categorized into 1000 classes. For data augmentation, we simply use random cropping and horizontal flipping. 

During the training of the masked transformer, a batch size of 512 was employed over 300 epochs, utilizing 8 GPUs (768 GPU hours on Nvidia A100) for 755,200 iterations. Subsequently, we fine-tuned the same model for $\sim$750,000 additional iterations on ImageNet $512\times 512$ with a batch size of 128 and 384 GPU hours on Nvidia A100. In total, this project, including training, testing and debugging, consumed $\sim$3,500 GPU hours on Nvidia A100.

While a lot of information to reproduce the results was available in the main paper, some points remain uncertain.
In particular, we discuss here the main differences between the paper, the official github implementation, and our own design choices:
\begin{itemize}
    \item For the architecture of the unmasking transformer, we try to stay as close as possible to the JAX implementation.
But it is still unclear why the class embeddings and the visual embeddings are shared in a single layer. This oddity results in implementing the token classification head as a dot product between the outputs of the transformer and each of the embeddings (classes and visual tokens), keeping only the similarity with visual embeddings to compute the cross-entropy.
The remaining 1001 values that correspond to the similarity with the class embeddings are dropped in the inference code, and we don't know if they were actually used for training in the official release.
    \item For sampling, the Gumbel noise injected in the confidence is not mentioned in the paper but taken from the official implementation.
    \item In addition, and not documented in the official releases, we used classifier-free guidance, and chose a learning rate of \(1e^{-4}\).
    \item We also use a different number of inference steps (15 steps) to achieve higher performance when compared to the original paper (12 steps).
\end{itemize}

\section{Performance on ImageNet}
\label{sec:results}

\subsection{Quantitative Results}
\begin{figure}
  \centering
  \begin{subfigure}[b]{0.48\textwidth}
    \centering
    \includegraphics[width=\textwidth]{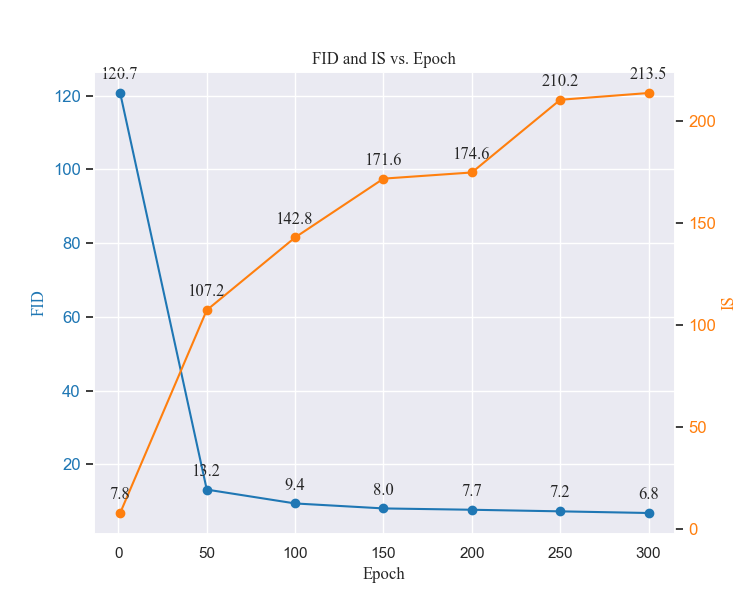}
    \caption{Training FID and IS (ImageNet $256\times 256$)}
    \label{fig:lr_curve}
  \end{subfigure}
  \hfill
  \begin{subfigure}[b]{0.48\textwidth}
    \centering
    \includegraphics[width=\textwidth]{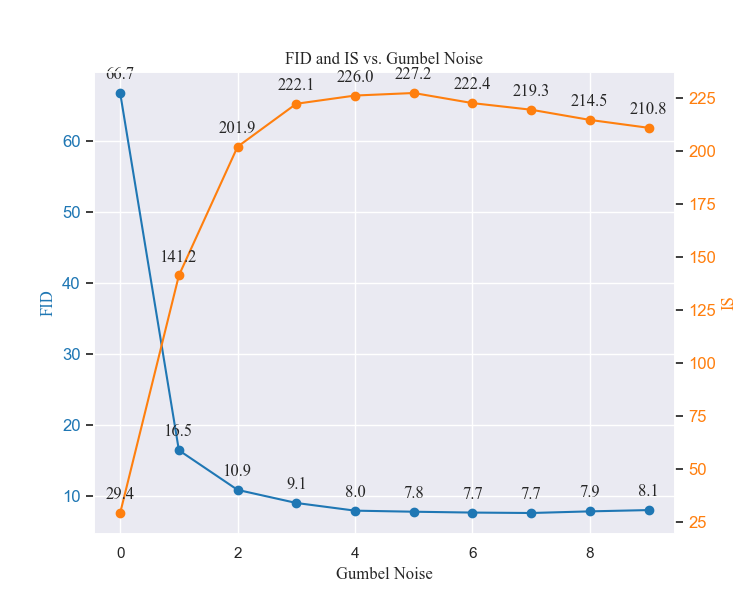}
    \caption{Gumbel noise (ImageNet $512\times 512$)}
    \label{fig:gumbel_curve}
  \end{subfigure}
  \\
  \begin{subfigure}[b]{0.48\textwidth}
    \centering
    \includegraphics[width=\textwidth]{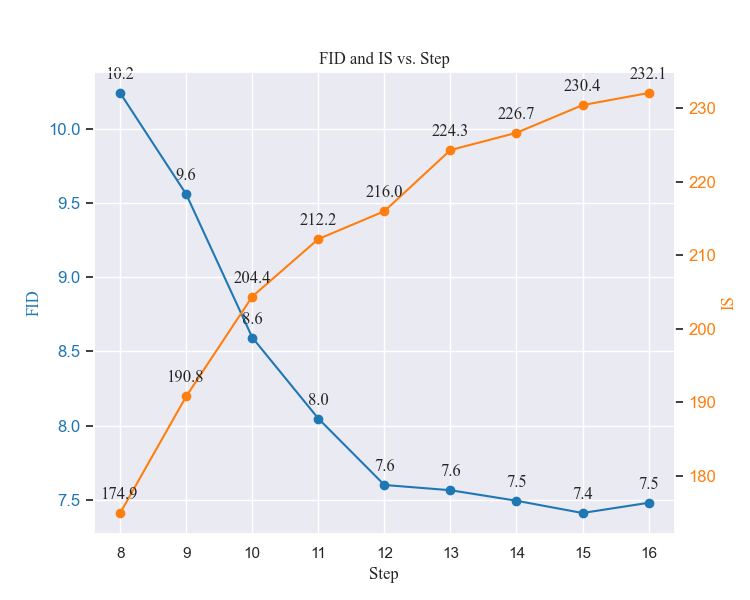}
    \caption{Number of steps (ImageNet $512\times 512$)}
    \label{fig:step}
  \end{subfigure}
  \hfill
  \begin{subfigure}[b]{0.48\textwidth}
    \centering
    \includegraphics[width=\textwidth]{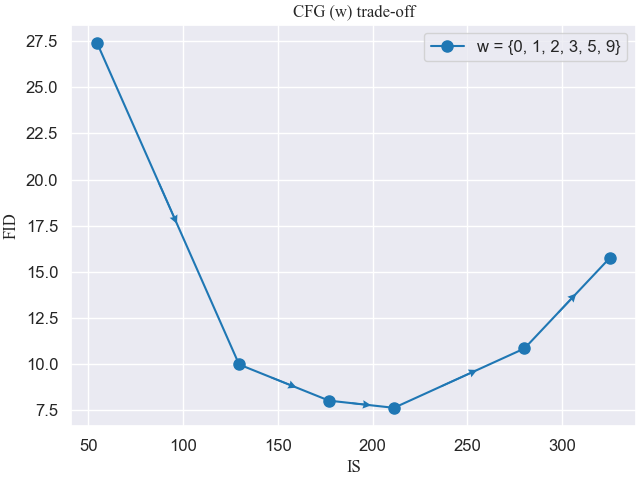}
    \caption{Classifier Free Guidance weight ($512\times 512$)}
    \label{fig:cfg}
  \end{subfigure}
  \caption{\textbf{Hyperparameters search:} ablation on crucial parameters to control sampling quality and diversity: number of training epochs, Gumbel noise, number of steps and the cfg weight.}
  \label{fig:full_curve}
\end{figure}

Using the hyperparameters for sampling presented in Table \ref{tab:hyperparameters}, we report our quantitative results in Table \ref{tab:metric_comparison}, as well as those from the original paper.
Overall, our reproduction shows very similar results for both ImageNet $256\times 256$ and $512\times 512$.
Slight differences are that our reproduction emphasizes better quality (Inception Score , Precision) and matches the diversity (Fréchet Inception Distance, Recall).
While only 8 steps are sufficient to generate good images in the lower resolution, one must increase the number of decoding steps up to 15 to achieve the best performance for ImageNet $512\times 512$. Moreover, a higher Gumbel noise is required to maintain diversity at higher resolutions (see also Figure \ref{fig:gumbel_curve}).
Finally, it is noteworthy that the best performance is achieved using the arccos scheduler, but results vary according to the number of steps, more information in Table \ref{tab:scheduler}.

\begin{table}
    \centering
    \caption{Hyperparameter used to get the best FID score.}
    \label{tab:hyperparameters}
    \begin{tabular}{cccccc} \toprule
        \textbf{Resolution} & \textbf{Softmax Temp.} & \textbf{Gumbel Temp.} & \textbf{CFG (w)} & \textbf{Schedule} & \textbf{Step} \\ \midrule
        $256\times 256$ & 1.0 & 4.5 & 3.0 & arccos & 8 \\
        $512\times 512$ & 1.0 & 7.0 & 2.8 & arccos & 15 \\ \bottomrule
    \end{tabular}
\end{table}

\begin{table}
\centering
    \caption{Comparison between this work (Ours) and the official paper on ImageNet 256 and 512.}
    \label{tab:metric_comparison}
    \begin{tabular}{l@{\hskip 0.5in} cccc} \toprule
        \textbf{Metric}                  & \textbf{Ours}      & \textbf{MaskGIT} \cite{chang2022maskgit}     & \textbf{Ours}      & \textbf{MaskGIT} \cite{chang2022maskgit}    \\ 
                                         & \textbf{($256\times 256$)} & \textbf{($256\times 256$)} & \textbf{($512\times 512$)} & \textbf{($512\times 512$)} \\ \midrule
        FID                              & 6.80               & \textbf{6.18}      & \textbf{7.26}      & 7.32               \\
        IS                               & \textbf{214.0}     & 182.1              & \textbf{223.0}     & 156.0              \\
        Precision                        & \textbf{0.82}      & 0.80               & \textbf{0.85}      & 0.78               \\
        Recall                           & \textbf{0.51}      & \textbf{0.51}      & 0.49               & \textbf{0.50}      \\
        Density                          & 1.25               & -                  & 1.33               & -                  \\
        Coverage                         & 0.84               & -                  & 0.86               & -                  \\
        \bottomrule
    \end{tabular}
\end{table}

In Figure \ref{fig:full_curve}, we present an ablation study on various hyperparameters related to the sampling process. For each curve or table representing an ablation study, all hyperparameters except the one being tested are held constant, as outlined below: scheduler is set to arccos, number of step is set to 12, softmax temperature is set to 1.0, cfg weight is set to 3.0 and Gumbel temperature: 7. 

Figure \ref{fig:lr_curve} show that the Inception Score (IS) exhibits a consistent increase throughout the training process, while the Fréchet Inception Distance (FID) demonstrates a corresponding decrease. This curve suggests that prolonging the training could potentially enhance the performance significantly, particularly for ImageNet $256\times 256$. 

Furthermore, we find out that injecting noise in the confidence score substantially improves the diversity of generated samples. Indeed, as can be seen in Figure \ref{fig:gumbel_curve}, introducing Gumbel noise in the confidence score, with a linear decay over the sampling, significantly improves the FID from $66.7$ to $7.7$. This trick, taken from the official code, was not discussed in the original paper.

Additionally, we study the influence of the number of steps on the quality of the samples. As expected, we show in Figure \ref{fig:step} that as the number of steps increases, the sampling quality improves. However, it is notable that the performance saturates beyond a certain number of steps, reaching best FID at 15 steps. 

Also, in Figure \ref{fig:cfg}, we perform experiments on the classifier free guidance (cfg) to strike a balance between fidelity, represented by FID, and quality, as indicated by IS.
An optimal trade-off seems to manifest at approximately cfg = 3, highlighting the importance of appropriate configuration for achieving the desired results with our model.

Finally, in Table \ref{tab:scheduler}, we conduct a comparative analysis of various schedulers at $512\times 512$ pixel resolution. Despite the model being trained with an arccos scheduler, the results indicate that other schedulers at inference could have some advantages. Indeed, when using 12 steps, square scheduler significantly improves FID while maintaining a reasonably high Inception Score. Conversely, a root scheduler yields superior quality, albeit at the cost of a higher FID. Unfortunately, none of these schedulers scale to a higher number of steps, and best results are obtained with the arccos scheduler with 15 steps or more. Nevertheless, these results imply the choice of a scheduler should change based on the number of sampling steps and compute budget.

\begin{table}
\centering
\caption{Scheduler configurations and metrics on ImageNet $512\times 512$}
\label{tab:scheduler}
\begin{tabular}{lccccccc}
\toprule
\textbf{Scheduler} & \textbf{FID} & \textbf{IS}      & \textbf{Precision} & \textbf{Recall} & \textbf{Density} & \textbf{Coverage} \\ \midrule
 root                   & 8.21          & \textbf{252.96} & \textbf{0.8619}    & 0.4644          & \textbf{1.3699} & \textbf{0.8666}    \\
 linear                 & 7.80          & 238.21          & 0.8547             & 0.4682          & 1.3505          & 0.8633             \\
 square                 & \textbf{7.50} & 224.07          & 0.8355             & \textbf{0.4904} & 1.2887          & 0.8524             \\
 cosine                 & 7.59          & 229.42          & 0.8423             & 0.4832          & 1.3052          & 0.8546             \\
 arccos                 & 7.65          & 219.32          & 0.8345             & 0.4826          & 1.2794          & 0.8437             \\
\bottomrule
\end{tabular}
\end{table}

\subsection{Qualitative Results}
\begin{figure}
  \centering
  \begin{subfigure}[b]{0.32\textwidth}
    \centering
    \includegraphics[width=\textwidth]{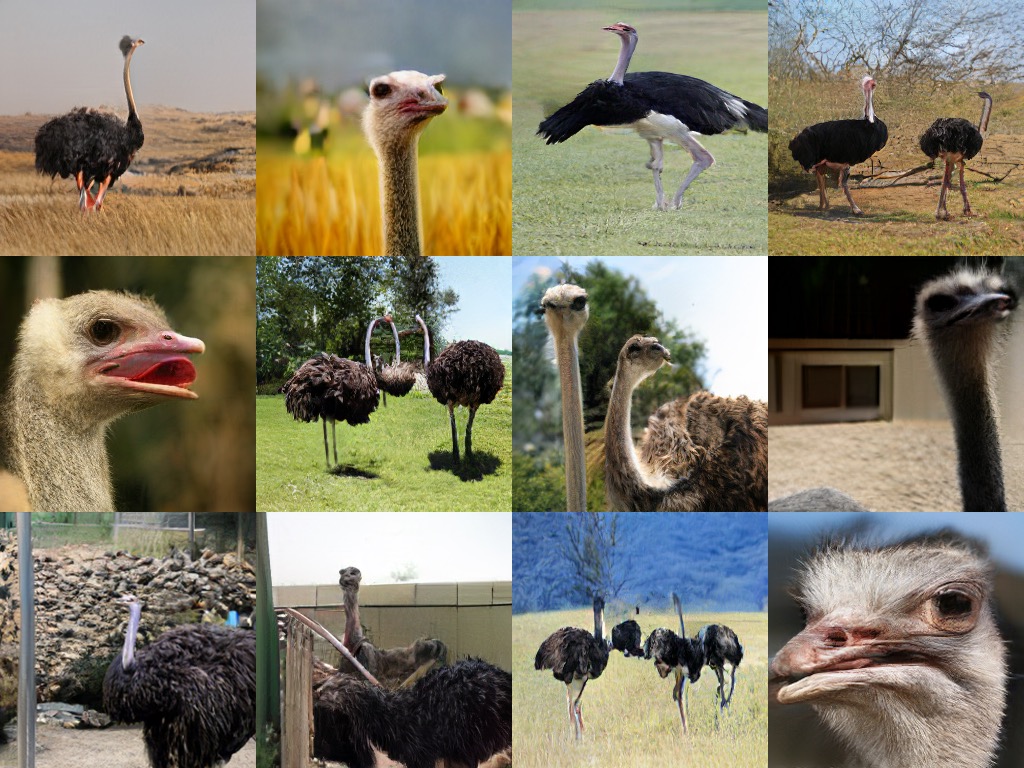}
    \caption{Official Ostrich (\imagenetid{009})}
    \label{fig:sub1}
  \end{subfigure}
  \hfill
  \begin{subfigure}[b]{0.32\textwidth}
    \centering
    \includegraphics[width=\textwidth]{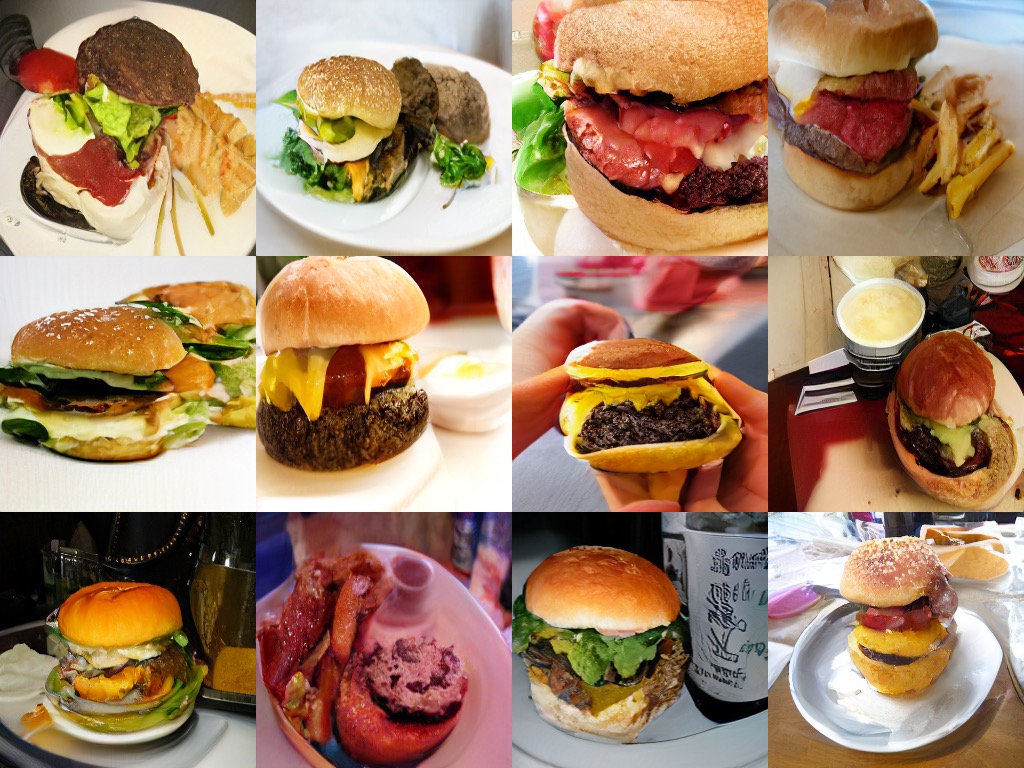}
    \caption{Official Burger (\imagenetid{933})}
    \label{fig:sub2}
  \end{subfigure}
  \hfill
  \begin{subfigure}[b]{0.32\textwidth}
    \centering
    \includegraphics[width=\textwidth]{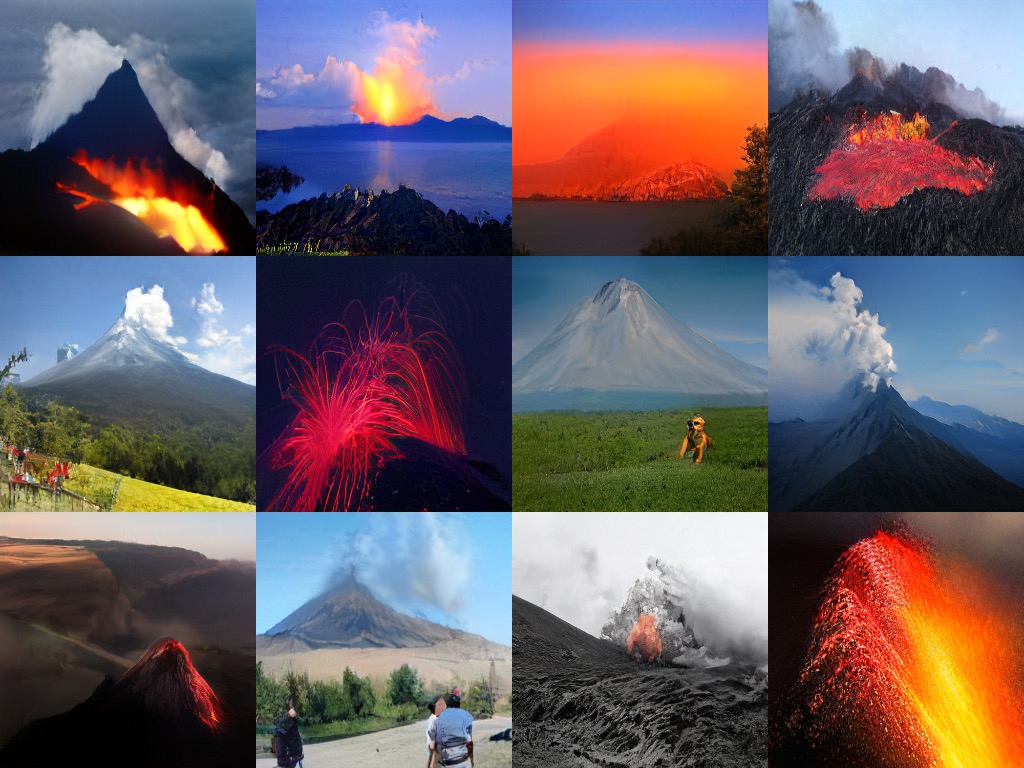}
    \caption{Official Volcano (\imagenetid{980})}
    \label{fig:sub3}
  \end{subfigure}
  \begin{subfigure}[b]{0.32\textwidth}
    \centering
    \includegraphics[width=\textwidth]{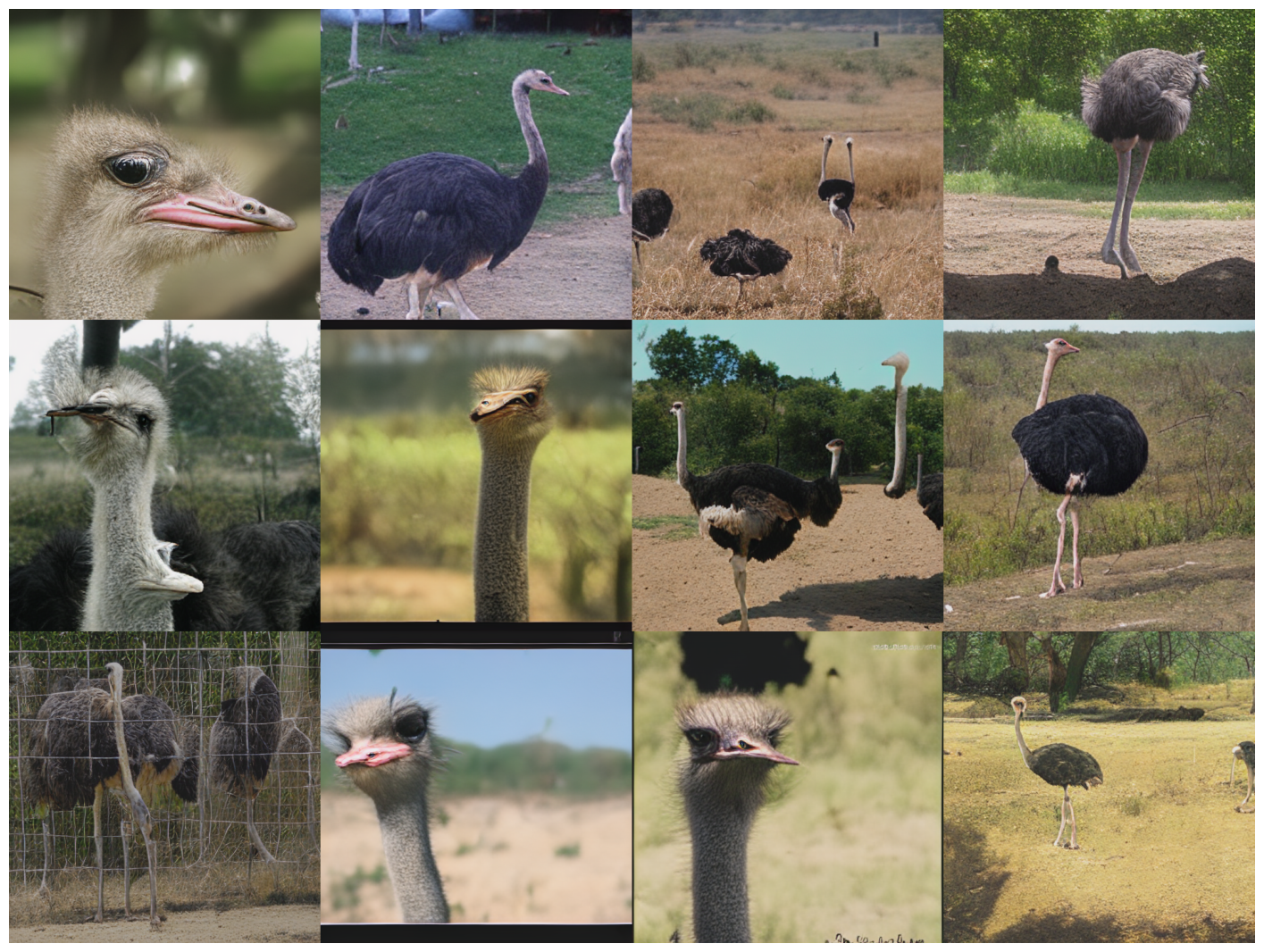}
    \caption{Ours Ostrich (\imagenetid{009})}
    \label{fig:sub4}
  \end{subfigure}
  \hfill
  \begin{subfigure}[b]{0.32\textwidth}
    \centering
    \includegraphics[width=\textwidth]{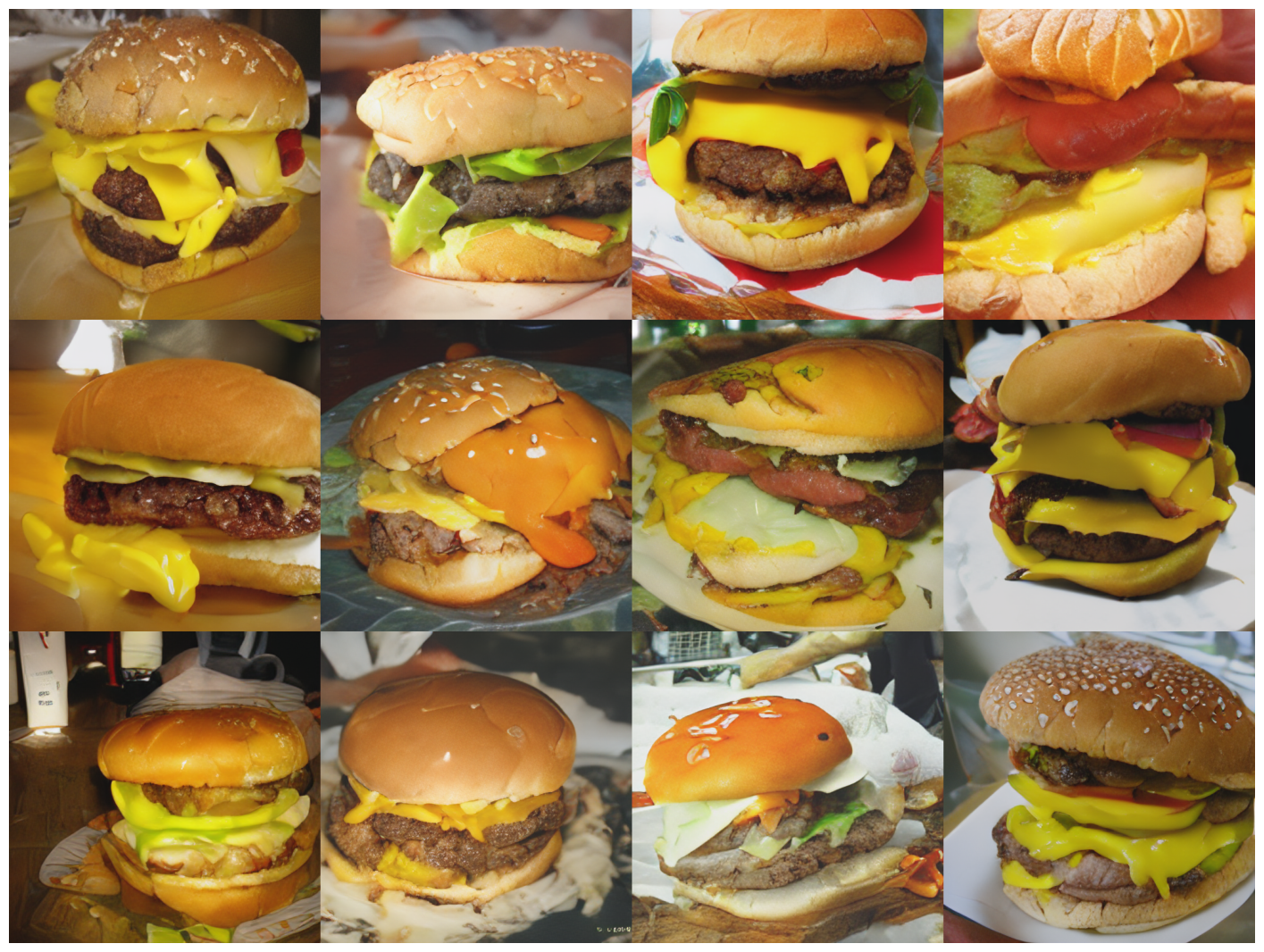}
    \caption{Ours Burger (\imagenetid{933})}
    \label{fig:sub5}
  \end{subfigure}
  \hfill
  \begin{subfigure}[b]{0.32\textwidth}
    \centering
    \includegraphics[width=\textwidth]{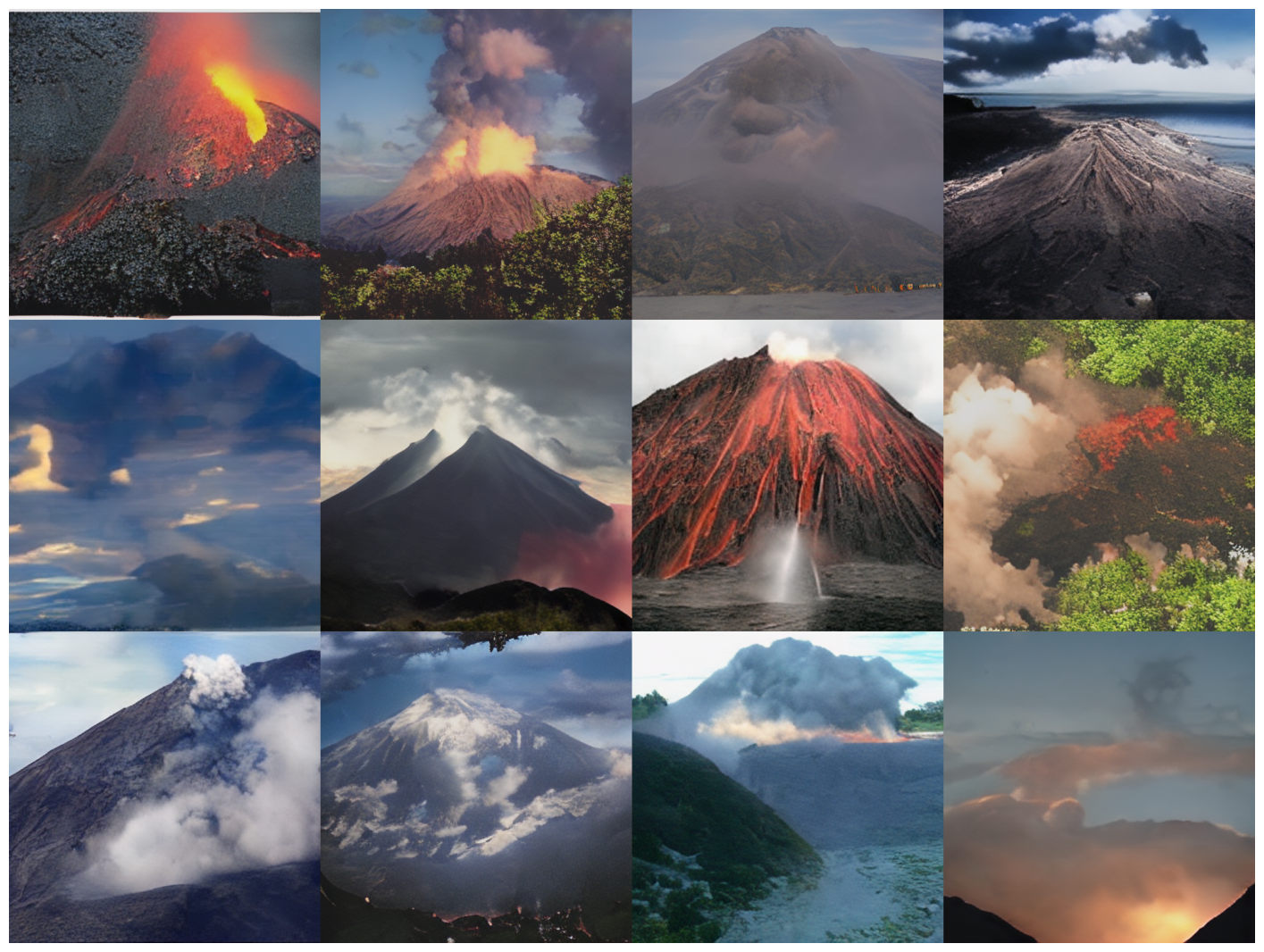}
    \caption{Ours Volcano (\imagenetid{980})}
    \label{fig:sub6}
  \end{subfigure}
  \caption{\textbf{Diversity comparison} between the official paper (top row) and our reproduction (bottom row). Without cherry pinking on our side, our methods exhibit a little bit less diversity but a higher quality}
  \label{fig:full_diversity}
\end{figure}

Here, we analyze our visual outcomes, showcased in Figure \ref{fig:qualitative_512}, focusing on $512\times 512$ resolution samples. These particular examples represent the visually best among the 10 random samples for each class. Each image requires 0.9365 seconds and 32 steps to be generated. In Figure \ref{fig:full_diversity}, we compare our samples to those showcased in the official paper. Our results, while seemingly displaying slightly less diversity, highlight higher quality, and are not cherry-picked. Moreover, those images take only 0.4406 seconds and 15 steps to be generated at $512\times 512$ resolution. And only 0.036 seconds is required to generate a $256\times 256$ pixels image with 8 forward steps. All results were obtained with Nvidia A100 GPUs. It is important to notice that it is $\sim$ 64x faster than auto-regressive methods.

\begin{figure}
    \centering    \includegraphics[width=0.9\linewidth]{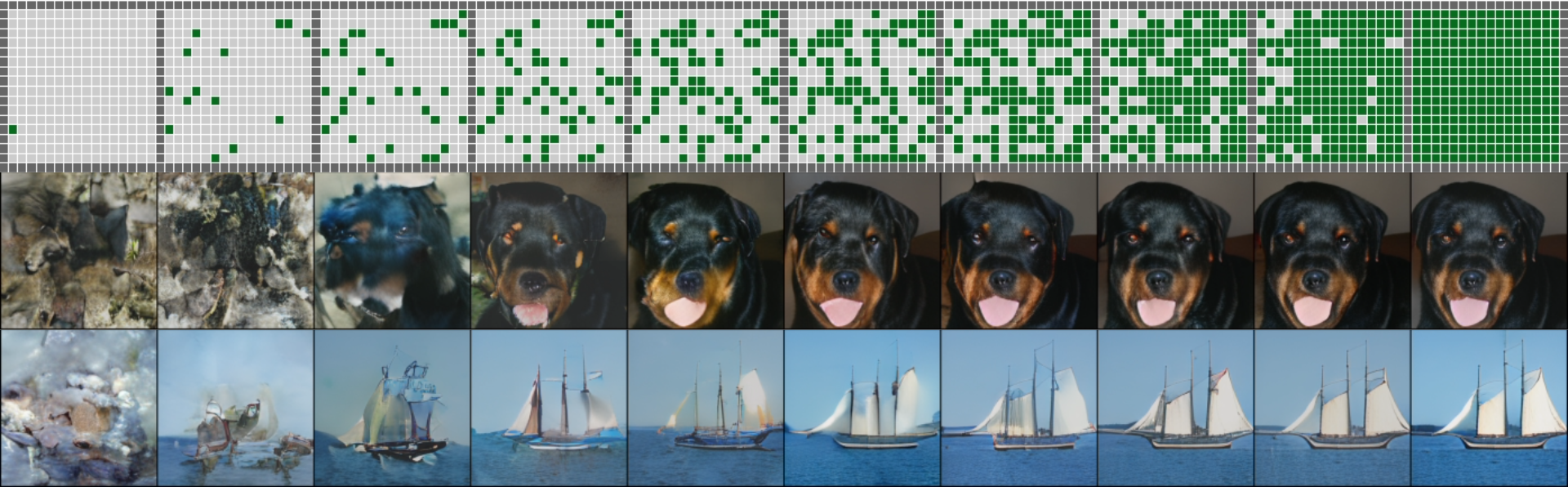} 
    \caption{\textbf{Intermediate images} generated at $256\times 256$ resolution. The first row showcases the binary mask, the second row exhibits the dog generated in association with the binary mask, and the last row presents another example of a sailboat.}
    \label{fig:intermediate_gen}
\end{figure}

In Figure \ref{fig:intermediate_gen}, we display the intermediate predictions of our models during the sampling process. Remarkably, the model produces reasonably good images even at an early stage of sampling.

\begin{figure}
    \centering
    \includegraphics[width=0.9\linewidth]{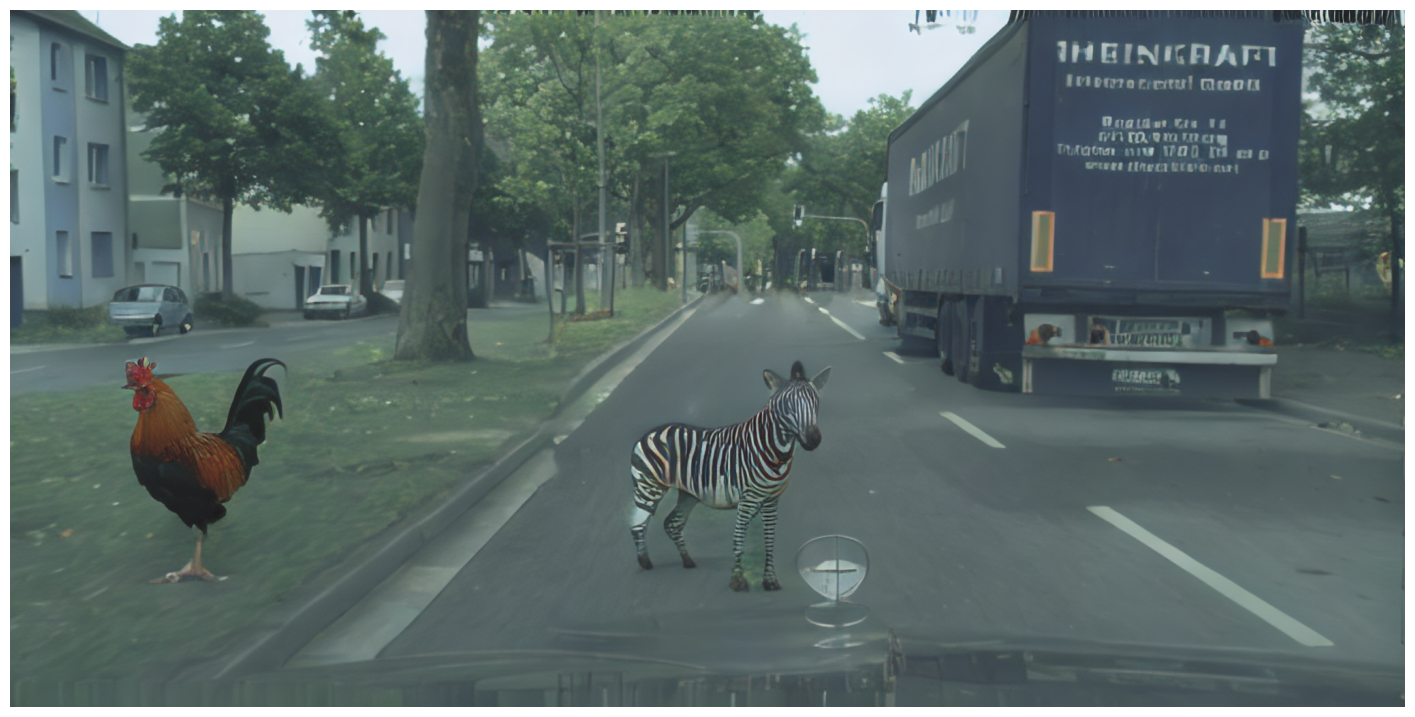} 
    \caption{\textbf{Image inpainting}:A rooster (\imagenetid{ImageNet 007}) and a zebra (\imagenetid{ImageNet 340}), generated and inpainted in a Cityscapes image.}
    \label{fig:inpainting}
\end{figure}

Additionally, Figure \ref{fig:inpainting} presents our inpainting results for a rooster (\imagenetid{ImageNet 007}) and a zebra (\imagenetid{ImageNet 340}) integrated into a Cityscapes image. We demonstrate that our adaptation of MaskGIT successfully achieves filling ImageNet classes within a road scene, showcasing the potential of our approach in this domain.

\section{Conclusion}
We released a Pytorch \cite{paszke2019pytorch} reproduction of MaskGIT \cite{chang2022maskgit}, the founding work for Masked Generative models.
Possible next direction is to extend the repository and add models for different modalities, like text-to-image (e.g. MUSE \cite{chang2023muse}) or text-to-video (e.g. MAGVIT \cite{yu2023magvit}).
It can also incorporate further Masked Generation improvement such as Token Critic \cite{lezama2022tokencritic} or Frequency Adaptive Sampling \cite{lee2023fas}.
It is our hope that this work can help other researchers to further explore the capabilities of Masked Generative Modeling.

\section*{Acknowledgments}

This research received the support of EXA4MIND, a European Union's Horizon Europe Research and Innovation program under grant agreement N°101092944. Views and opinions expressed are however those of the author(s) only and do not necessarily reflect those of the European Union or the European Commission. Neither the European Union nor the granting authority can be held responsible for them.

\bibliographystyle{plain}
\bibliography{ref}

\end{document}